\documentclass[10pt, a4paper]{article}
\usepackage{lrec2022} 
\usepackage{multibib}
\newcites{languageresource}{Language Resources}
\usepackage{graphicx}
\usepackage{tabularx}
\usepackage{color}
\usepackage{soul}
\usepackage{multirow}

\usepackage{titlesec}
\titleformat{\section}{\normalfont\large\bfseries\center}{\thesection.}{1em}{}
\titleformat{\subsection}{\normalfont\SmallTitleFont\bfseries\raggedright}{\thesubsection.}{1em}{}
\titleformat{\subsubsection}{\normalfont\normalsize\bfseries\raggedright}{\thesubsubsection.}{1em}{}
\renewcommand\thesection{\arabic{section}}
\renewcommand\thesubsection{\thesection.\arabic{subsection}}
\renewcommand\thesubsubsection{\thesubsection.\arabic{subsubsection}}

\usepackage{epstopdf}
\usepackage[utf8]{inputenc}

\usepackage{hyperref}
\usepackage{xstring}
\usepackage{cleveref}
\usepackage{colortbl}

\title{Slovene SuperGLUE Benchmark: Translation and Evaluation}

\name{Aleš Žagar, Marko Robnik-Šikonja} 

\address{
University of Ljubljana, Faculty of Computer and Information Science,  Večna pot 113, 1000 Ljubljana,  \\        ales.zagar@fri.uni-lj.si, marko.robnik@fri.uni-lj.si\\}

\abstract{
We present a Slovene combined machine-human translated SuperGLUE benchmark. We describe the translation process and problems arising due to differences in morphology and grammar. We evaluate the translated datasets in several modes: monolingual, cross-lingual, and multilingual, taking into account differences between machine and human translated training sets. The results show that the monolingual Slovene SloBERTa model is superior to massively multilingual and trilingual BERT models, but these also show a good cross-lingual performance on certain tasks. The performance of Slovene models still lags behind the best English models. 
 \\ \newline  }

\begin{document}

\maketitleabstract

\section{Introduction}
To measure the progress across the fast evolving area of natural language processing (NLP), several prominent benchmarking suites have been proposed, such as SentEval \cite{Conneau2018}, GLUE \cite{wang2019glue}, and SuperGLUE \cite{wang2019superglue}.
SentEval aims to evaluate sentence encoders, while GLUE and its more challenging successor, SuperGLUE, evaluate general language understanding. These two benchmark suites contain public training sets and private test sets that can be assessed through an evaluation server giving separate scores for each of the tasks and an overall single-number score. The tasks in SuperGLUE are diverse and comprised of question answering (QA), natural language inference (NLI), coreference resolution, and word sense disambiguation (WSD). Non-expert humans evaluated all the tasks to provide a human performance baseline. 

Despite recent criticism of the way how the evaluation scores in these benchmarks are computed (e.g., the arithmetic mean of separate metrics is used for all tasks of different complexity and different sizes of testing sets) \cite{shavrina2021hownottolie,kasai2021bidimensional}, there is little doubt that the datasets contained in the benchmarks significantly contribute towards the progress of the NLP area. Unfortunately, most of the research is on English, which often limits the generality of the approaches and does not address the full scale of language complexity. Still, we can observe significant efforts to make relevant datasets cross-lingual or adapted to more languages. Broad multilingual benchmarking suites, such as XTREME \cite{hu2020xtreme} and XGLUE \cite{liang2020xglue}, provide machine-translated datasets for several relevant tasks and languages. Complementary to that, we present an in the direction of less-resourced languages, namely a combined machine-human translation of SuperGLUE benchmarking suite to Slovene. We describe the difficulties arising in the translation and provide the first evaluation of the datasets using large pretrained BERT-like language models. Our evaluation encompasses monolingual, cross-lingual and multilingual approaches, comparison of human and machine-translated parts, and comparison of two state-of-the-art machine translation systems.

The paper is structured into further five sections. In \Cref{sec:related}, we present related work on non-English benchmarking suites. In \Cref{sec:benchmark}, we describe the translation process and the final SuperGLUE datasets in Slovene. We describe three evaluation settings (monolingual, cross-lingual, and multilingual) in \Cref{sec:evaluation}, and present the results in \Cref{sec:results}. \Cref{sec:conclusions} presents conclusions, limitations of our work, and ideas for further improvements. To make the paper self-contained and ease understanding of the different datasets in the SuperGLUE benchmark, we follow \newcite{wang2019superglue} and present an example for each of the tasks in Appendix.

\section{Related work}
\label{sec:related}
In the area of NLP, there are several prominent benchmarking suites, the most prominent being GLUE \cite{wang2019glue} and SuperGLUE \cite{wang2019superglue}. While most benchmarking datasets are available in English, there are a few notable attempts to extend them to other languages.

 XTREME \cite{hu2020xtreme} is  a multi-task benchmark for evaluating the cross-lingual representations across 40 languages and 9 tasks. It focuses on the zero-shot cross-lingual transfer and provides large English training sets, much smaller testing sets in target languages (from 5 to 40 human translated or checked, the rest are machine translated without human interventions). The suite covers tasks such as natural language inference (NLI), paraphrasing, part-of-speech (POS) tagging, named entity recognition (NER), question answering, and parallel sentence extraction and detection.  

 XGLUE \cite{liang2020xglue} is intended to evaluate cross-lingual pretrained models on 11 tasks: NER, POS-tagging, question answering, paraphrasing, ad relevance, web page relevance, question-answer matching, question generation, and news title generation. Similarly to XTREME, the suite provides training data in English and tests the cross-lingual models on testing data on several target languages (from 3 to 18).
 
 SuperGLUE is likely the most prominent benchmarking suite for English. It follows the design of the GLUE suite, consisting of a public leaderboard for eight tasks, which are evaluated individually and jointly. The suite provides public training and development datasets, while testing data is hidden and only used to evaluate predictions submitted to the leaderboard. The benchmark includes QA, NLI, coreference resolution, and WSD tasks, for which non-expert human baselines are provided.
 
 Following the English example, Russian SuperGLUE \cite{shavrina2020russiansuperglue} was developed independently or manually translated from English SuperGLUE. It contains nine tasks: linguistic diagnostic, WSD, QA, NLI, and coreference resolution.
 
 Slovene language possesses significantly fewer resources than English or Russian and is not included in the XTREME and XGLUE massively cross-lingual benchmarks. We aimed to provide a modern natural language understanding suite for this less-resourced language using modest resources at our disposal. For that purpose, we followed the English SuperGLUE design and translated the entire datasets (except WiC) to Slovene, mainly using machine translation and in small part human translation (120,000 words). This allows testing of monolingual, cross-lingual and multilingual approaches. Keeping the format of the original SuperGLUE, our datasets can be evaluated with the original SuperGLUE leaderboard and can be compared to English baselines and state-of-the-art. 
 
\section{Slovene SuperGLUE translation}
\label{sec:benchmark}

To evaluate cross-lingual transfer and test specifics of morphologically rich languages, we translated the SuperGLUE datasets to Slovene. Due to limited funds, we partially used human translation (HT) and partially machine translation (MT). Altogether, approximately 120,000 words were human translated. Some datasets are too large (BoolQ, MultiRC, ReCoRD, RTE) to be fully human translated. We thus provide ratios between the human translated and the original English sizes in Table \ref{tab:superglue_translation}. For MT from English to Slovene, we used the GoogleMT Cloud service. In our evaluation, we use six of the original eight tasks. As explained below, we excluded ReCoRD and WiC.

\begin{table}[htb]
\caption{Number of instances in the original English and translated Slovene SuperGLUE tasks. HT stands for human translation and MT for machine translation. The ``ratio`` indicates the ratio between the number of human translated instances and all instances. }
\label{tab:superglue_translation}
\centering
\resizebox{\columnwidth}{!}{
\begin{tabular}{l|lllll}
\textbf{Dataset} & \textbf{split} & \textbf{English} & \textbf{HT} & \textbf{ratio} & \textbf{MT} \\ \hline
\textbf{BoolQ} & train & 9427 & 92 & 0.0098 & yes \\
\textbf{} & val & 3270 & 18 & 0.0055 & yes \\
\textbf{} & test & 3245 & 30 & 0.0092 & yes \\ \hline
\textbf{CB} & train & 250 & 250 & 1.0000 & yes \\
\textbf{} & val & 56 & 56 & 1.0000 & yes \\
\textbf{} & test & 250 & 250 & 1.0000 & yes \\ \hline
\textbf{COPA} & train & 400 & 400 & 1.0000 & yes \\
\textbf{} & val & 100 & 100 & 1.0000 & yes \\
\textbf{} & test & 500 & 500 & 1.0000 & yes \\ \hline
\textbf{MultiRC} & train & 5100 & 15 & 0.0029 & yes \\
\textbf{} & val & 953 & 3 & 0.0031 & yes \\
\textbf{} & test & 1800 & 30 & 0.0167 & yes \\ \hline
\textbf{ReCoRD} & train & 101000 & 60 & 0.0006 & / \\
\textbf{} & val & 10000 & 6 & 0.0006 & / \\
\textbf{} & test & 10000 & 30 & 0.0030 & / \\ \hline
\textbf{RTE} & train & 2500 & 232 & 0.0928 & yes \\
\textbf{} & val & 278 & 29 & 0.1043 & yes \\
\textbf{} & test & 300 & 29 & 0.0967 & yes \\ \hline
\textbf{WiC} & train & 6000 & / & / & / \\
\textbf{} & val & 638 & / & / & / \\
\textbf{} & test & 1400 & / & / & / \\ \hline
\textbf{WSC} & train & 554 & 554 & 1.0000 & / \\
\textbf{} & val & 104 & 104 & 1.0000 & / \\
\textbf{} & test & 146 & 146 & 1.0000 & / \\ \hline
\end{tabular}
}
\end{table}

We decided to use only the HT translated part of our test sets in our evaluations to avoid noise due to translations. This makes some test sets much smaller compared to English test sets. We did not include ReCoRD in the Slovene benchmark due to the low quality of our test set, consisting of confusing and ambiguous examples. Further, there are differences between English and Slovene ReCoRD tasks due to the morphological richness of Slovene. Namely, in Slovene, the correct declension of a query is often not present in the text, making it impossible to provide the right answer. Finally, similarly to WSC (discussed below), ReCoRD is also affected by the problem of translating HTML tags with GoogleMT. 

The WiC task cannot be translated and would have to be conceived anew because it is impossible to transfer the same set of meanings of a given word from English to a target language. Taking the WiC example in Table \ref{tab:superglue_examples}, we note that the word \emph{board} in two different contexts translates to two completely different words in Slovene (Context 1: Bivanje in \emph{hrana}. Context 2: Čez okna je nabil \emph{deske}.).  

The WSC dataset cannot be machine-translated because it requires human assistance and verification. First, GoogleMT translations cannot handle the correct placement of HTML tags indicating coreferences. 
The second reason is that in Slovene coreferences can also be expressed with verbs, while coreferences in English are mainly nouns, proper names and pronouns. 
This makes the task different in Slovene compared to English. On the one hand, the task is more difficult in Slovene because solutions cover more types of words; on the other hand, the Slovene verbs might reveal the coreference information for some instances.

\section{Evaluation settings}
\label{sec:evaluation}
SuperGLUE benchmark is extensively used to compare large pretrained models in English\footnote{\href{https://super.gluebenchmark.com/leaderboard}{https://super.gluebenchmark.com/leaderboard}}. In contrast to that, we concentrate on the Slovene translation of the SuperGLUE tasks and new opportunities which arise from that. 

We compare four BERT models available for Slovene: monolingual Slovene SloBERTa \cite{ulcar2021sloberta}, trilingual (Croatian-Slovene-English) CroSloEngual BERT \cite{ulcar2020xlbert}, massively multilingual mBERT \cite{Devlin2019} (bert-base-multilingual-cased\footnote{\href{https://huggingface.co/bert-base-multilingual-cased}{https://huggingface.co/bert-base-multilingual-cased}}), and XLM-R \cite{conneau2019unsupervised} (xlm-roberta-base\footnote{\href{https://huggingface.co/xlm-roberta-base}{https://huggingface.co/xlm-roberta-base}}).  

The single-number overall average score (i.e. Avg in the second column of, e.g., Table \ref{tab:superglue_results}) comprises equally weighted tasks. For English, there are eight tasks; for Slovene, there are all six  translated tasks: BoolQ, CB, COPA, MultiRC, RTE, and WSC. In tasks with multiple metrics, we averaged those metrics to get a single task score. For the details on how the score is calculated for each task, see \cite{wang2019superglue}.

We test models in three settings. The monolingual setting uses the same language (Slovene or English) for training and testing. In the cross-lingual setting, we tested the cross-lingual models (CroSloEngual, mBERT, and XLM-R) and transfer between English and Slovene datasets in both directions. In the multilingual setting, the models were trained on the combined full size English and Slovene data. The results for the three settings as well as a comparison of two MT systems (GoogleMT and DeepL) are reported in \Cref{sec:results}.

We fine-tuned the available Slovene BERT models on SuperGLUE tasks using the Jiant tool \cite{phang2020jiant}. We used a single-task learning setting for each task and fine-tuned models for 100 epochs, with the initial learning rate of $10^{-5}$. Each model was fine-tuned using the dataset corresponding to one of the three settings. 

\section{Results}
\label{sec:results}
We first report the results for each of the three settings (monolingual, cross-lingual, and multilingual) separately. We end the section with the comparison between human and machine-translated data and the comparison of machine translation systems.

\begin{table*}[!!ht]
\caption{The SuperGLUE benchmarks in English (upper part) and Slovene (lower part). All English results are taken from \protect \cite{wang2019superglue}. 
The best score for each task and language is in \textbf{bold}. The best average scores (Avg) for each language are \underline{underlined}. *MultiRC has multiple test sets released on a staggered schedule, and these results evaluate on an installation of the test set that is a subset of ours.}
\label{tab:superglue_results}
\resizebox{\textwidth}{!}{%
\begin{tabular}{l|lllllllll}
\textbf{Task} & \textbf{Avg} & \textbf{BoolQ} & \textbf{CB} & \textbf{COPA} & \textbf{MultiRC} & \textbf{ReCoRD} & \textbf{RTE} & \textbf{WiC} & \textbf{WSC} \\
\textbf{Models/Metrics} &  & \textbf{Acc.} & \textbf{F1/Acc.} & \textbf{Acc.} & \textbf{F1$_a$/EM} & \textbf{F1/EM} & \textbf{Acc.} & \textbf{Acc.} & \textbf{Acc.} \\ \hline
Most Frequent  & 45.7 & 62.3 & 21.7/48.4 & 50.0 & 61.1/0.3 & 33.4/32.5 & 50.3 & 50.0 & \textbf{65.1} \\
CBoW  & 44.7 & 62.1 & 49.0/71.2 & 51.6 & 0.0/0.4 & 14.0/13.6 & 49.7 & 53.0 & \textbf{65.1} \\
BERT  & 69.3 & 77.4 & 75.7/83.6 & 70.6 & 70.0/24.0 & \textbf{72.0/71.3} & 71.6 & \textbf{69.5} & 64.3 \\
BERT++  & \underline{73.3} & \textbf{79.0} & \textbf{84.7/90.4} & \textbf{73.8} & \textbf{70.0/24.1} & \textbf{72.0/71.3} & \textbf{79.0} & \textbf{69.5} & 64.3 \\ Human (est.)  & 89.8 & 89.0 & 95.8/98.9 & 100.0 & 81.8*/51.9* & 91.7/91.3 & 93.6 & 80.0 & 100.0 \\ \hline 

Most Frequent &  49.1 & 63.3 & 21.7/48.4 & 50.0 & \textbf{76.4/0.6} & - & 58.6 & - & 65.8  \\ 

SloBERTa       & \underline{63.9} & 66.6     &  \textbf{74.0/76.8} & \textbf{61.8}   & 62.7/21.9    & -    & \textbf{62.1} & - & \textbf{73.3}    \\
CroSloEngual   & 57.8 & 66.6     &  62.1/72.4 & 58.2   & 56.7/15.6    & -    & \textbf{62.1} & - & 56.2    \\
mBERT          & 59.1 & 70.0       &  66.6/73.6 & 54.2   & 57.4/16.3    & -    & \textbf{62.1} & - & 61.6    \\
XML-R          & 58.7 & \textbf{76.7}     &  66.2/73.2 & 50.0   & 55.3/13.9    & -    & 55.2 & - & 65.8    \\ \hline
\end{tabular}%
}
\end{table*}

\subsection{Monolingual results}
Our monolingual analysis compares different Slovene prediction models on the complete datasets, composed of existing human translated instances while the remaining instances are machine-translated.  
Table \ref{tab:superglue_results} shows the results together with several baselines trained on the original English datasets. Some comparisons to English baselines are questionable as the Slovene models are trained on only a small fraction of better quality HT data (BoolQ, MultiRC) and tested on smaller set of HT test data. In the case of the BERT++ model, the English model was additionally pretrained with transfer learning tasks similar to a target one (CB, RTE, BoolQ, COPA). In terms of datasets, the only fair comparison is possible with the CB, COPA, and WSC.  

Considering the Avg scores in Table \ref{tab:superglue_results}, the monolingual SloBERTa is the best performing Slovene model. On average, all Slovene BERT models perform better than the Most Frequent baseline. Concerning individual tasks, none of the Slovene models exceeds the Most Frequent baseline in the MultiRC task. SloBERTa was significantly better than the rest of the models in CB, COPA, and WSC, while XML-R was the best on BoolQ. 

Compared to English models, the best Slovene model (SloBERTa) achieved better results on WSC. It seems that none of the English models learned anything from WSC (they are below the Most Frequent baseline), but the SloBERTa model achieved the score of 73.3 (the Most Frequent baseline gives 65.8). The success of SloBERTa on WSC might stem from the morphology of Slovene verbs, which include the information on the gender; this information is beneficial in coreference resolution and makes some instances easier in Slovene than in English. Nevertheless, there is still a large gap compared to human performance. All Slovene models showed good performance on CB and fell between English CBoW and BERT. 




\subsection{Cross-lingual results}
In the cross-lingual scenario, we tested the three multilingual BERT models (mBERT, CroSloEngual, XLM-R) and the transfer between English and Slovene datasets (both directions). For Slovene as the source language, we used the available human translated examples. To make the comparison balanced, we only used the same examples from the English datasets. We tested both zero-shot transfer (no training data in the target language) and few-shot transfer. In the few-shot training, we used 10 additional examples from the target language for each task. We randomly sampled these 10 examples for 5 times and reported averages to achieve more statistically valid results. The fine-tuning hyperparameters are the same as in the monolingual setup. 

\begin{table*}[!!ht]
\centering
\caption{Cross-lingual results on human translated SuperGLUE test sets. The best results for zero-shot and few-shot scenarios are in bold. }
\label{tab:cross-lingual_superglue}
\resizebox{\textwidth}{!}{%
\begin{tabular}{l|l|ll|lllllll}
\multirow{2}{*}{\textbf{Evaluation}} & \multirow{2}{*}{\textbf{Model}} & \multirow{2}{*}{\textbf{source}} & \multirow{2}{*}{\textbf{target}} & \textbf{Avg} & \textbf{BoolQ} & \textbf{CB} & \textbf{COPA} & \textbf{MultiRC} & \textbf{RTE} & \textbf{WSC} \\
 &  &  &  &  & \textbf{acc.} & \textbf{F1/acc.} & \textbf{Acc.} & \textbf{F1$_a$/EM} & \textbf{Acc.} & \textbf{Acc.} \\ \hline
\multirow{6}{*}{Zero-shot} & 

 \multirow{2}{*}{CroSloEngual} & english & slovene & 49.8 & 56.7 & 43.7/60.0 & 54.6 & 48.0/6.6 & 58.6 & 50.7 \\
 &  & slovene & english & 52.6 & 60.0 & 53.8/70 & \textbf{59.6} & \textbf{56.7/9.6} & 48.3 & 58.2 \\
 
 & \multirow{2}{*}{mBERT} & english & slovene & 47.4 & 56.7 & 36.2/57.2 & 50.2 & 47.3/8.7 & 55.2 & 64.4 \\
 &  & slovene & english & 48.3 & 60.0 & 44.6/50.4 & 49.8 & 56.2/8.7 & 51.7 & 57.5 \\ 
 
  & \multirow{2}{*}{XLM-R} & english & slovene & \textbf{53.8} & \textbf{63.3} & \textbf{62.9/68.4} & 53.6 & 48.5/0.3 & \textbf{62.1} & 56.2 \\
 & & slovene & english & 51.7 & \textbf{63.3} & 59.1/67.2 & 47.2 & 52.9/12.9 & 51.7 & \textbf{65.8} \\ \hline
 
\multirow{6}{*}{Few-shot} & 

\multirow{2}{*}{CroSloEngual} & english & slovene & 54.4 & 60.0 & 52.4/68.6 & 55.0 & 52.8/9.72 & \textbf{65.5} & 54.1 \\
 &  & slovene & english & 53.0& 60.0	& 53.8/70.0 &	\textbf{59.5} &	56.0/12.1 &	49.7 &   58.2 \\

 & \multirow{2}{*}{mBERT} & english & slovene & 50.9 & 60.1 & 53.1/66.2 & 50.4 & 50.8/9.8 & 53.8 & 64.4 \\
 &  & slovene & english & 51.3 & 60.7 &	51.8/58.2 &	50.3 &	57.2/11.1&	56.5&	56.8
 \\
 
  & \multirow{2}{*}{XLM-R} & english & slovene & \textbf{57.0} & \textbf{63.3} & \textbf{65.8/69.8} & 53.3 & \textbf{76.4/0.6} & 62.1 & 57.4 \\
 &  & slovene & english & 53.0 & \textbf{63.3} &	63.0/69.6&	48.3 &	51.4/10.6	& 55.8 &	\textbf{65.8}
 \\ \hline
 
 & Most frequent &  &  & 52.4 & 63.3 & 23.0/52.7 & 50.0 & 77.3/0.3 & 58.6 & \textbf{65.8}
\end{tabular}
}
\end{table*}

\begin{table*}[!!ht]
\caption{Multilingual results on English and Slovene (human translated) test sets. Models were trained on combined full size English and Slovene data. Slovene data is comprised of human translated and machine translated data. The best results for each task is in bold.  }
\label{tab:multilingual_results}
\centering
\begin{tabular}{l|l|lllllll}
\textbf{Evaluation on}      & \textbf{Model} & \textbf{Avg}  & \textbf{BoolQ} & \textbf{CB}        & \textbf{COPA} & \textbf{MultiRC}   & \textbf{RTE}  & \textbf{WSC}  \\
\textbf{}                & \textbf{}      & \textbf{}     & \textbf{Acc.}  & \textbf{F1/acc.}   & \textbf{Acc}  & \textbf{F1a/EM}    & \textbf{Acc.} & \textbf{Acc.} \\ \hline
\multirow{3}{*}{Slovene} & CroSloEngual   & 59.8          & 70.0           & 67.7/74.7          & 59.4          & 58.4/15.6          & 51.7          & 58.2          \\
                         & mBERT          & 60.2          & 73.0           & 66.5/71.9          & 51.6          & 57.5/17.0          & 62.1          & 58.9          \\
                         & XML-R          & 59.9          & 63.3           & 69.9/74.7          & 52.8          & 58.8/18.8          & 58.6          & 61.0          \\ \hline
\multirow{3}{*}{English} & CroSloEngual   & 59.9          & 63.3           & 67.3/75.5          & \textbf{62.4} & 59.5/16.7          & 55.2          & 57.5          \\
                         & mBERT          & \textbf{64.2} & \textbf{76.7}  & 69.9/74.7          & 58.6          & \textbf{60.4/21.5} & \textbf{65.5} & 63.0          \\
                         & XML-R          & 61.4          & 70.0           & \textbf{74.1/79.9} & 51.8          & 60.1/19.4          & 48.3          & \textbf{65.8} \\ \hline
                         & Most frequent  & 52.4          & 63.3           & 23.0/52.7        & 50            & 77.3/0.3         & 58.6          & 65.8         
\end{tabular}
\end{table*}

\begin{table*}[tb]
\caption{Comparing human translation (HT) with machine translation (MT) on the Slovene SuperGLUE benchmark. Note that for this experiment the average score Avg is computed as the average of the five listed tasks and not six as in \Cref{tab:superglue_results}. The best score for each task is in \textbf{bold}.}
\label{tab:HTandMT}
\centering
\begin{tabular}{l|llllll}
\textbf{Task} & \textbf{Avg} & \textbf{BoolQ} & \textbf{CB} & \textbf{COPA} & \textbf{MultiRC} &  \textbf{RTE}  \\
\textbf{Models/Metrics} &  & \textbf{Acc.} & \textbf{F1/Acc.} & \textbf{Acc.} & \textbf{F1$_a$/EM} &  \textbf{Acc.}  \\ \hline
Most Frequent &  49.1 & 63.3 & 21.7/48.4 & 50.0 & \textbf{76.4/0.6} &  \textbf{58.6}  \\ 
HT-mBERT                &  54.3 & 63.3 & 66.6/73.6 & 54.2 & 45.1/8.1 &  57.2   \\ 
MT-mBERT                &  55.2 & 63.3 & 65.1/68.8 & 54.4 & 55.4/11.7 &  57.9   \\ 
HT-CroSloEngual         &  55.6 & 63.3 & 62.1/72.4 & 58.2 & 53.0/8.4 & \textbf{58.6}  \\ 
MT-CroSloEngual         &  53.4 & 63.3 & 59.8/68.4 & 55.0 & 51.2/10.5 & 53.8   \\ 
HT-SloBERTa             &  \textbf{57.2} & 63.3 & \textbf{74.0/76.8} & \textbf{61.8} & 53.0/10.8 & 53.8  \\ 
MT-SloBERTa             &  55.8 & 63.3 & 68.6/74.8 & 58.2 & 57.1/12.0 &  49.6   \\ 
HT-XLM-R          &  53.5 & 63.3 & 66.2/73.2 & 50.0 & 53.3/0.9 &  57.2  \\ 
MT-XLM-R          &  50.1 & 63.3 & 62.0/68.4 & 51.4 & 55.3/0.6 &  42.8  \\ \hline

HT-Avg & \textbf{55.1} & 63.3 & \textbf{70.6} & \textbf{56.0} & 29.1 &  \textbf{56.7} \\ 
MT-Avg & 53.6 & 63.3 & 67.0 & 54.8 & \textbf{31.7} &  51.0 \\ \hline

\end{tabular}%
\end{table*}

The results are presented in Table \ref{tab:cross-lingual_superglue}. Averaged over all tasks, some models improved the Most frequent baseline. In general, they were quite unsuccessful on BoolQ, MultiRC, and WSC but showed some promising results on COPA, RTE, and especially CB. Additional training examples in the few-shot scenario brought some visible improvements. It seems that models perform better in the English-Slovene direction than vice versa. The best performing model is XLM-R, followed by CroSloEngual BERT and mBERT. 

The low overall performance can be explained by a low number of training examples in the source language. If we take a closer look at specific models, we can observe that XLM-R shows very good results on CB in both directions. CroSloEngual BERT achieved a similar outstanding result on COPA. It is the only model that learned something on this dataset as well.    

We can conclude that for the difficult SuperGLUE benchmark, the cross-lingual transfer is challenging but not impossible. In the future, we plan to expand the current set of experiments in several directions. First, we will train English models on the complete SuperGLUE datasets and transfer them to Slovene human and machine-translated datasets. Second, we will train Slovene models on the combined machine and human translated datasets and transfer them to complete English datasets. 
Finally, we will also combine training for several tasks and test transfer learning scenarios.

\subsection{Multilingual results}
In the multilingual setting, the three multilingual BERT models (CroSloEngual BERT, mBERT, and XLM-R) were trained on the combined full-size English and Slovene data. Slovene data is comprised of human-translated and machine-translated data. For Slovene, the models are tested on only HT data. The results are reported in 
\Cref{tab:multilingual_results}.

Interestingly, all the best scores for all tasks were achieved when tested on English testing sets (the training sets were identical and comprised of both languages). This might be due to the fact that Slovene BERT models were pretrained on lower amounts of data and the quality of the Slovene translation.  

The overall best model with the highest Avg score is mBERT (best in BoolQ, MultiRC and RTE), followed by XLM-R (best in CB and WSC) and CroSloEngual BERT (best in COPA). In the MultiRC and WSC all models lag behind the Most Frequent baseline.


\subsection{Comparing human and machine translation}
To test the difference in human and machine translation, we repeated the monolingual experiments separately on human translated (HT) data and the same machine translated (MT) instances (the sizes of these HT datasets are reported in \Cref{tab:superglue_translation}). Each model was fine-tuned using either MT or HT datasets of the same size. Only the translated content varies between both translation types; otherwise, they contain exactly the same examples. The splits of instances into train, validation and test sets is the same as in the English variant  (but mostly considerably smaller, see Table \ref{tab:superglue_translation}). WSC is excluded from this evaluation as it can only be human translated, so the average score (Avg) is computed from the five remaining tasks. Table \ref{tab:HTandMT} shows the results. The comparison with results in \Cref{tab:superglue_results} is not fair because the models there used significantly more training examples. 

Considering the Avg scores in Table \ref{tab:HTandMT}, the monolingual SloBERTa is again the best performing Slovene model and all BERT models, regardless of translation type, perform better than the Most Frequent baselines. From the translation type perspective, the models trained on HT datasets perform better than those trained on MT datasets by 1.5 points. 
The only task where MT is better than HT is MultiRC, but looking at single scores, we can observe that none of the models learned anything in this task as there is a large gap between the Most Frequent baseline and the rest of the models. 

Analysis of specific tasks shows that 
For the BoolQ dataset, all models predict the most frequent class (the testing set might be too small for reliable conclusions in BoolQ). We can safely assume that training sample sizes are too small in BoolQ and MultiRC (none of the models learned anything) and must be increased (we have only 92 HT examples in BoolQ and 15 HT examples in MultiRC). The same is also true for RTE.

\subsection{Comparing machine translation systems}
To check the effect of MT systems on the performance of models, we perform a separate study in the monolingual setting using two currently best MT systems for Slovene: GoogleMT\footnote{\href{https://translate.google.com/}{https://translate.google.com/}} and DeepL\footnote{\href{https://www.deepl.com/translator}{https://www.deepl.com/translator}}. 

\begin{table}[htb]
\centering
\caption{Comparison of GoogleMT and DeepL MT systems on the CB and COPA datasets. The best score for each task is in bold.}
\label{tab:mt_systems}
\resizebox{\columnwidth}{!}{%
\begin{tabular}{l|cc|cc}
 & \multicolumn{2}{c|}{\textbf{GoogleMT}} & \multicolumn{2}{c}{\textbf{DeepL}} \\
\textbf{Model} & \textbf{CB} & \textbf{COPA} & \textbf{CB} & \textbf{COPA} \\ \hline
SloBERTa & 68.6/74.8 & \textbf{58.2} & \textbf{73.6/79.1} & 54.2 \\
CroSloEngual & 59.8/68.4 & 55.0 & 66.1/75.1 & 53.8 \\
mBERT & 65.1/68.8 & 54.4 & 60.4/70.7 & 50.8 \\
XML-R & 62.0/68.4 & 51.4 & 44.3/63.5 & 53.6 \\ \hline
\end{tabular}
}
\end{table}

We compare all four Slovene BERT models on two datasets where all models performed better than the Most Frequent baseline, CB and COPA (WSC needs manual translation as discussed in \Cref{sec:benchmark}). Results in \Cref{tab:mt_systems} show surprisingly large differences between the two MT systems. DeepL gives favourable results on the CB dataset, while the inverse is true on the CB dataset. We cannot explain the differences and leave a more detailed analysis of the issue for further work.

\section{Conclusions}
\label{sec:conclusions}
We prepared the Slovene translation of natural language understanding benchmark suite SuperGLUE and released it under an open-source licence \citelanguageresource{SloSuperGLUE}\footnote{\href{http://hdl.handle.net/11356/1380}{http://hdl.handle.net/11356/1380}}. 
We described the translation process and obstacles in the transfer to a morphologically rich language. The partially machine and partially human translated datasets were used in the assessment of four BERT-based models available for Slovene. The results show that the monolingual SloBERTa model is currently the best performing Slovene pretrained model. However, the performance of Slovene models is still significantly worse compared to the state-of-the-art English models, showing considerable potential for improvement of NLP approaches for less-resourced languages.

Our analyses show that the models' performance improved with human translated datasets, and in future, we intend to increase the share of human translated data. For two of the English SuperGLUE datasets, the MT was not possible. We intend to create a Slovene version of the WiC task from scratch using word sense disambiguation tasks and manually adapt the ReCoRD task to get the full SuperGLUE suite. In the WSC tasks, Slovene verb morphology leaks some coreference information. We intend to analyze this issue and form a more challenging Slovene WSC.   
Next, both Slovene and Russian are Slavic languages, and therefore more similar to each other than English. It would be interesting to combine equivalent Russian and Slovene tasks from the translated SuperGLUE tasks in further multilingual and cross-lingual experiments.
Finally, as we kept the format of the original SuperGLUE benchmarks, Slovene datasets can be evaluated with the original SuperGLUE leaderboard. While this allows comparison to English models, the obtained results are not publicly available. We intend to prepare a separate Slovene leaderboard and encourage the NLP community to pay attention to less-resourced languages.

\section*{Acknowledgements}
The work was partially supported by the Slovenian Research Agency (ARRS) core research programmes P6-0411, as well as the Ministry of Culture of Republic of Slovenia through project Development of Slovene in Digital Environment (RSDO).
This paper is supported by European Union's Horizon 2020 research and innovation programme under grant agreement No 825153, project EMBEDDIA (Cross-Lingual Embeddings for Less-Represented Languages in European News Media).
We thank the authors of SuperGLUE benchmark for sharing test set answers for some of the tasks.

\onecolumn
\section*{Appendix}
\begin{table*}[ht]
\footnotesize
\centering
\caption{Examples from the development set of SuperGLUE tasks. \textbf{Bold} texts represent parts of examples' format. Texts in \textit{italics} are part of models' input. \underline{Underlined} texts are specially
marked in inputs. Texts in the \texttt{monospaced font} represent the expected models' outputs. The WiC task cannot be translated into other languages as different meanings of a single  word would translate into different words. The ReCoRD cannot be handled with MT systems as they do not preserve tagged words. Further, in morphologically rich languages such as Slovene, the correct answer would not necessarily use the same word form as the one tagged.}
\begin{tabular}{p{0.02\linewidth}p{0.85\linewidth}}
\hline
\multirow{3}{*}{\rotatebox[origin=c]{90}{\textbf{BoolQ}}} & \textbf{Passage}:\textit{Barq's -- Barq's is an American soft drink. Its brand of root beer is notable for having caffeine. Barq's, created by Edward Barq and bottled since the turn of the 20th century, is owned by the Barq family but bottled by the Coca-Cola Company. It was known as Barq's Famous Olde Tyme Root Beer until 2012.} \\
& \textbf{Question}:is barq's root beer a pepsi product \\
& \textbf{Answer}:\texttt{No} \\ \hline

\multirow{3}{*}{\rotatebox[origin=c]{90}{\textbf{CB}}} & \textbf{Text}:\textit{B: And yet, uh, I we-, I hope to see employer based, you know, helping out. You know, child, uh, care centers at the place of employment and things like that, that will help out. A: Uh-huh. B: What do you think, do you think we are, setting a trend?
} \\ 
& \textbf{Hypothesis}:they are setting a trend \\
& \textbf{Entailment}:\texttt{Unknown} \\ \hline

\multirow{5}{*}{\rotatebox[origin=c]{90}{\textbf{COPA}}} & \textbf{Premise}:\textit{My body cast a shadow over the grass.
} \\ 
& \textbf{Question}:What's the CAUSE for this \\
& \textbf{Alternative 1}:The sun was rising. \\ 
& \textbf{Alternative 2}:The grass was cut. \\
& \textbf{Correct Alternative}:\texttt{1} \\ \hline

\multirow{3}{*}{\rotatebox[origin=c]{90}{\textbf{MultiRC}}} & \textbf{Paragraph}:\textit{Susan wanted to have a birthday party. She called all of her friends. She has five friends. Her mom said that Susan can invite them all to the party. Her first friend could not go to the party because she was sick. Her second friend was going out of town. Her third friend was not so sure if her parents would let her. The fourth friend said maybe. The fifth friend could go to the party for sure. Susan was a little sad. On the day of the party, all five friends showed up. Each friend had a present for Susan. Susan was happy and sent each friend a thank you card the next week. 
} \\ 
& \textbf{Question}:Did Susan's sick friend recover? \\
& \textbf{Candidate answers}: Yes, she recoverd, No (\texttt{F}), Yes (\texttt{T}), No, she didn't recover (\texttt{F}), Yes, she was at Susan's party (\texttt{T}) \\ \hline

\multirow{3}{*}{\rotatebox[origin=c]{90}{\textbf{ReCoRD}}} & \textbf{Paragraph}:\textit{(\underline{CNN}) \underline{Puerto Rico} on Sunday overwhelmingly voted for statehood. But Congress, the only body that can approve new states, will ultimately decide whether the status of the \underline{US} commonwealth changes. Ninety-seven percent of the votes in the nonbinding referendum favored statehood, an increase over the results of a 2012 referendum, official results from the \underline{State Electoral Commission} show. It was the fifth such vote on statehood. Today, we the people of \underline{Puerto Rico} are sending a strong and clear message to the \underline{US Congress} ... and to the world ... claiming our equal rights as \underline{American} citizens, \underline{Puerto Rico} Gov. \underline{Ricardo Rossello} said in a news release.@highlight \underline{Puerto Rico} voted Sunday in favor of \underline{US} statehood
} \\ 
& \textbf{Query}:For one, they can truthfully say, ''Don't blame me, I didn't vote for them,'' when discussing the $\langle$placeholder$\rangle$ presidency. \\
& \textbf{Correct Entities}: \texttt{US} \\ \hline

\multirow{3}{*}{\rotatebox[origin=c]{90}{\textbf{RTE}}} & \textbf{Text}:\textit{Dana Reeve, the widow of the actor Christopher Reeve, has died of lung cancer at age 44, according to the Christopher Reeve Foundation. 
} \\ 
& \textbf{Hypothesis}:Christopher Reeve had an accident. \\
& \textbf{Entailment}: \texttt{False} \\ \hline

\multirow{3}{*}{\rotatebox[origin=c]{90}{\textbf{WiC}}} & \textbf{Context 1}:\textit{Room and \underline{board}.  
} \\ 
& \textbf{Context 2}:He nailed \underline{boards} across the windows. \\
& \textbf{Sense match}: \texttt{False} \\ \hline

\multirow{2}{*}{\rotatebox[origin=c]{90}{\textbf{WSC}}} & \textbf{Text}:\textit{Mark told \underline{Pete} many lies about himself, which Pete included in his book. \underline{He} should have been more truthful.  
} \\ 
& \textbf{Coreference}: \texttt{False} \\ \hline

\end{tabular}

\label{tab:superglue_examples}
\end{table*}

\twocolumn
\clearpage
\section{Bibliographical References}\label{reference}
\label{main:ref}

\bibliographystyle{lrec2022-bib}
\bibliography{NLP,embeddia-refs}

\section{Language Resource References}
\label{lr:ref}
\bibliographystylelanguageresource{lrec2022-bib}
\bibliographylanguageresource{languageresource}

\end{document}